\RequirePackage{fix-cm}
\documentclass{svjour3} % onecolumn (standard format)
\smartqed % flush right qed marks, e.g. at end of proof
\usepackage{subcaption}
\usepackage{graphicx, multicol, latexsym, amsmath, amssymb,xcolor}
\usepackage{blindtext}
\usepackage{textcomp}
\usepackage{subcaption}
\usepackage[sort,numbers]{natbib}
\usepackage{url}

\begin{document}

 \title{A New GNG Graph-Based Hand Gesture Recognition Approach}
%\subtitle{Do you have a subtitle?\\ If so, write it here}

 %\titlerunning{Short form of title} % if too long for running head

 \author{Narges Mirehi \and
Maryam Tahmasbi
}
%\authorrunning{Short form of author list} % if too long for running head
\institute{ Maryam Tahmasbi \at
Department of computer science, Shahid Beheshti University, G.C., Tehran, Iran.\\
Tel.: +98 21 2990 3004\\
Fax: +98 21 2243 1653\\
\email{m\_tahmasbi@sbu.ac.ir} \\
%\emph{Present address:} of F. Author % if needed
\and
Narges Mirehi
\at
Department of computer science, Shahid Beheshti University, G.C., Tehran, Iran.\\
\email{n\_mirehi@sbu.ac.ir}\\
}
%\authorrunning{Short form of author list} % if too long for running head

%\date{Received: date / Accepted: date}
% The correct dates will be entered by the editor

\maketitle

 \begin{abstract}
Hand Gesture Recognition (HGR) is of major importance for Human-Computer Interaction (HCI) applications.
In this paper, we present a new hand gesture recognition approach called GNG-IEMD. In this approach, first, we use a Growing Neural Gas (GNG) graph to model the image. Then we extract features from this graph. These features are not geometric or pixel based, so do not depend on scale, rotation, and articulation.
The dissimilarity between hand gestures is measured with a novel Improved Earth Mover\textquotesingle s Distance (IEMD) metric. We evaluate the performance of the proposed approach on challenging public datasets including NTU Hand Digits, HKU, HKU multi-angle, and
UESTC-ASL and compare the results with state-of-the-art approaches. The experimental results demonstrate the performance of the proposed approach.

 % \PACS{PACS code1 \and PACS code2 \and more}
% \subclass{MSC code1 \and MSC code2 \and more}
\end{abstract}

 \section{Introduction}
Nonverbal communication, which involves communication through hand gestures, body, and facial movements, contains about 65\% of all human communication \cite{Hogan2003}. Hand gestures are the most important part of nonverbal communication among body, arm, and facial movements (body language). One of the objectives of intelligent systems is to facilitate natural human-computer interaction.
Among various behaviors of human-computer interaction, hand gesture is a natural and effective way for communication with significant ability to exchange information.

 However, hand gesture recognition is known as a difficult problem in computer vision due to the varieties in the shape, size, and direction of hands or fingers in different hand images. The problem can be generally divided into two categories: static and dynamic conditions.
The recognition of dynamic gestures tries to examine spatial-temporal characteristics, while static detection focuses on the internal information of an image. The study of static gesture recognition is an essential part of hand gesture recognition because hand shapes carry specific information without any movement \cite{Li2018Deep}. %In this paper, we study static hand gesture recognition by GNG graph viewpoints.

 Most previous approaches have considered developing systems for hand gesture recognition using a combination of preprocessing and machine learning methods. Most of these studies extract pixel-based features and classify hand gestures using machine learning methods
\cite{dong2015american,Li2018Deep}.
The study of hand gesture recognition using meaningful shape features is important because meaningful shape features improve stability against articulation, scale, rotation, and noise.
Hence, in recent decades, researchers have been trying to recognize hand gestures using significant features extracted from the shape of the image and its boundary \cite{ren2011robust,wang2017Liu}. Skeleton, geometry and graph-based methods are the most well-known methods in this area \cite{ren2011robust,ren2013robust,wang2013hand,wang2017Liu}.\\
Skeleton-based methods have attractive properties such as invariance to scale and rotation by capturing topological and geometrical information of skeletal branches. The main limitation of skeleton-based methods is low stability against contour noises.
In fact, small variation or noise on the boundary of the object can cause redundant branches in the skeleton and significant changes in the structure of its topology.\\
Geometric-based methods are studied geometric conditions of the image, such as Euclidian distance and angle of fingers with respect to the center of the palm
and describe the important information of the object with a summary vector and ignore the redundant information of the pixels. These approaches may be influenced by articulation and viewpoint.
Most of them are based on hand contour, but they are often distorted due to low resolution and precision of the current depth cameras. In other words, their performance may be reduced due to orientation and noises on contour \cite{wang2017Liu}.

 Graphs are robust against rotation, articulation, and noise. The inherent properties of a graph do not depend on its representation, so it can be used as effective tools for image representation.
Hence, various graph-based methods have been presented for hand gesture recognition, but their structure depends on the local information of the pixels, and the loss of some pixel information, such as noise and small inner holes reduces their performance \cite{li2014hegm,triesch2002classification,wang2015superpixel,wang2017Liu}.
%Another significant algorithm introduced recently based on Growing Neural Gas graph (GNG) and Linear Discriminant Analysis (LDA) for hand gesture recognition \cite{mirehi2019hand}.
%In this paper, the proposed algorithm \cite{} extends
Recently, a new graph-based method that uses GNG to construct the graph and linear discriminant analysis (LDA) is introduced for hand gesture recognition\cite{mirehi2019hand}.
%In this paper, we aim to provide a new graph-based method for hand gesture recognition with less dependency on pixels compared to other existing approaches. We extend previous work \cite{mirehi2019hand} and present a graph-based algorithm with high ability to capture the topological and geometrical properties of a hand image.
%We use Growing Neural Gas (GNG) algorithm to model hand image which provides a graph whose vertices are distributed almost uniformly inside the object. Afterwards, we define features captures the shape of graph.
% We also introduce an improved version of Earth Mover\textquotesingle s Distance (EMD) to measure dissimilarity between feature vectors and evaluate the proposed method on challenging datasets including NTU Hand Digit dataset, HKU, HKU multi-angle, and UESTC-ASL dataset.
%The results indicate the high accuracy of the proposed method.

 In this paper, we use an idea similar to \cite{mirehi2019hand} to form a GNG graph for a given image then, we introduce new topological features for hand gesture recognition. The new features can extract convexity and concavity of boundaries more precisely. We also introduce an improved version of Earth Mover\textquotesingle s Distance to measure the dissimilarity between feature vectors.
This leads to higher accuracy in different datasets. We evaluate the proposed approach on challenging datasets including NTU Hand Digit dataset, HKU, HKU multi-angle, and UESTC-ASL dataset.% The results indicate the high accuracy of the proposed method.

The rest of the paper is organized as follows: Section \ref{sec:related works} reviews the related work briefly. Section \ref{sec:Proposed method} presents the basic steps of the proposed method and hand gesture recognition approach. The results of the experimental study and a comparison with state-of-the-art approaches are presented in Section \ref{sec:Experimental study}. Finally, Section \ref{sec:conclusion} concludes this paper.
\section{Related works} \label{sec:related works}
In this section, we review the state-of-art approaches briefly.
%Many vision-based hand gesture recognition algorithms have been proposed in the past years and comprehensive reviews can be found in
Different skin color methods have been used for hand detection and segmentation. The main decision on providing a skin color model is the choice of color space. However, variations of skin colors and background objects with color distribution similar to human skin can confuse the methods \cite{rautaray2015vision}.
Most of the current methods use Kinect sensors to collect data, detect and segment hand information \cite{Maqueda2015,cheng2016image,wang2017Liu,Li2018Deep}.
Depth cameras facilitate the hand segmentation process compared to skin-based models, especially when there are similar texture backgrounds \cite{plouffe2016static,ren2011robust,ren2013robust}. In these approaches, the hand of the user is considered as the nearest object of the scene to the camera and segmentation is performed by specifying a threshold value. We use the same way for hand detection and segmentation in this study.

 For more precision in hand detection and segmentation, some approaches applied both the depth map and skeleton tracking provided by Kinect for hand detection \cite{presti20163d,zafrulla2011american,wang2015superpixel}.
Although these methods may provide more accuracy in hand prediction, they suffer from configuration complexity.

 Various hand features can be used for hand recognition. Hand features can be grouped into almost two groups, including pixel-based \cite{Maqueda2015,zhang2013histogram} and shape-based features \cite{bai2008path,belongie2002shape,Stergiopoulou}.
Shape-based features contain geometry, graph, and skeleton-based features.

 Belongie et al. introduced a shape context descriptor by computing a log-polar histogram of the relative position of contour points \cite{belongie2002shape}.
Fritzke et al. presented an incremental network which learns the topological structure of input vectors by a simple Hebb-like rule \cite{Fritzke}.
\\
Stergiopoulou and Papamarkos applied GNG graph for image representation and considered limited geometric features such as the
distance and angle between neurons \cite{Stergiopoulou}.

 The skeleton of objects can be considered another source of shape information for hand gesture recognition \cite{bai2008path}. Noisy and distorted contours have a significantly negative effect on extraction of the correct skeleton.
Zhang et al. used local features for hand gesture recognition. They computed the Histogram of Oriented Gradients (HOG) of 3D point distribution in color images \cite{zhang2013histogram}.

 In \cite{Maqueda2015}, a Volumetric Spatiograms of Local Binary Patterns (VS-LBP) method was employed for hand gesture recognition.
Despite the appropriateness of the results, these approaches depend on the local information of the pixels, which reduces their stability against pixel distortions \cite{zhang2013histogram,Maqueda2015}.
Ren et al. in \cite{ren2011robust} and \cite{ren2013robust} proposed a contour-based method by a Finger Earth Mover\textquotesingle s Distance (FEMD) and a template matching approach. Contours are often distorted due to low resolution and the precision of the current depth cameras, which affects the accuracy of contour-based approaches.
Wang et al. presented a color-depth Superpixel Graph Earth Mover\textquotesingle s Distance (SP-EMD) constructed by segmenting pixels to almost the same size superpixels. They applied Earth Mover\textquotesingle s Distance (EMD) to measure the similarity of hand gestures and considered the cost of the centroid of superpixels based on their depth information and location, of them which can be influenced by camera conditions and the variety of hand shapes \cite{wang2015superpixel}.
Wang e.al extended the previous method based on Canonical Superpixel-Graph to reduce hand the shape variation problem \cite{wang2017Liu}.
In another study, a super-pixel based finger Earth Mover\textquotesingle s Distance (SPFEMD) approach was proposed which was considered only super-pixel of fingers and was used template matching \cite{wang2019hand}.
An Image-to-Class Dynamic Time Warping (I2C-DTW) approach for both 3D static and trajectory hand gesture recognition was introduced in \cite{cheng2016image} by computing the Image-to-Class distance for hand gesture classification.

 In \cite{plouffe2016static}, a K-curvature algorithm, which is based on the change in the slope angle of the tangent line, was employed for the localization the fingertips over the contour extracted from depth data, and dynamic time warping (DTW) is applied for gesture recognition. This approach depends on the precision and resolution of the depth data.

 Moreover, various deep learning approaches have been proposed for developing hand gesture recognition systems \cite{cheok2019review,farooq2015survey,Li2018Deep,nunez2018convolutional}.
Li et al. provided a deep CNN framework for hand gesture recognition using the four-channel RGB-D (Depth) of the image \cite{Li2018Deep}. Their disadvantage is their dependence on lighting conditions. Nunez et al. proposed the combination of a CNN and a Long-Short Term Memory (LSTM) network based on human skeleton kinematics for the hand gesture recognition problem \cite{nunez2018convolutional}.

 \section{Proposed approach}\label{sec:Proposed method}
In the previous work \cite{mirehi2019hand}, Growing Neural Gas (GNG) graphs were constructed from binary images and the bulges of hand including fingers and wrist were computed, afterward, topological and geometrical features were extracted from the GNG graph. The hand gestures were first categorized based on the number of bulges and then classified according to defined features by LDA.
In the current work, we extend method \cite{mirehi2019hand} in the following directions.

 1) Defined features are developed to recognize the hand gesture, and new features describing the shape of the boundary, such as concavity, convexity, and overall bulge shape are defined.

 2) A new Earth Mover\textquotesingle s Distance (EMD) is introduced to measure the dissimilarity of feature vectors extracted GNG graphs.

 3) Hand gestures are classified by k-NN classifier according to Earth Mover\textquotesingle s Distance.

 4) The proposed approach is evaluated on more challenging datasets including HKU, HKU multi-angle, and UESTC-ASL.
\\
We first briefly describe the overall framework and then describe each step in detail.
The main steps of the proposed approach contain:\\
1. Segmentation of the hand gesture image.\\
We use the simple thresholding on the depth map for hand segmentation.
We consider the hand of the user as the nearest object of the scene to the camera and segment the hand region by thresholding on the depth.
This is a simple and effective method for hand segmentation, which is applied in many approaches \cite{ren2011robust,ren2013robust}.\\
2. Constructing the GNG graph of the binary image. \\
3. Computing the outer boundary of the graph by computational geometry approaches. \\
4. Extracting the topological and geometrical features of the graph.\\
5. Measuring the dissimilarity between the hand gestures using a new Earth Mover\textquotesingle s Distance.\\
6. Classifying the hand gestures by k-NN algorithm.

 %\subsection{Segmentation}
%Depth cameras facilitate the hand segmentation process compared to skin-based models especially when there are similar texture backgrounds \cite{plouffe2016static,ren2011robust,ren2013robust}. In these approaches, the hand of the user is considered as the nearest object of the scene to the camera and the segmentation is performed by specifying a threshold value.

 \subsection{Constructing the GNG graph}
Various approaches can be applied to construct a graph for an image. Our target graph should provide the following properties:
\begin{itemize}
\item
The vertices should be distributed almost uniformly within the image, and the edges should be almost equal.
\item
The number of vertices should be constant in other words, the graph should not depend on the scale of the image.
\item
The graph should ignore the holes and cracks inside the image and be robust against the noise on the image contour.
\end{itemize}
We choose the GNG graph because it can very well satisfy the explained properties. The GNG algorithm includes a low-dimensional subspace of the input data space while learning the topological structure of data distribution \cite{Fritzke}. Moreover, the GNG algorithm has the ability to follow the behavior of vertices in changing dynamic conditions and can be extended to 3D online representation and object tracking \cite{fink2015novelty,orts20163d,sun2017online}.
In the following, we will describe the GNG algorithm briefly.
\subsection{The GNG graph}
Growing Neural Gas is an unsupervised learning algorithm \cite{Fritzke}. The algorithm starts with two vertices that are located in a random location and then updates the location of some vertices by comparing the distance of vertices with the initial data in each step.
Some error is assigned to the closest vertex to the compared input data, which indicates the distance between them.
The closest vertices to the compared input data are connected by a zero-age edge and old edges are removed from the graph.
Eventually, new vertices are added between the vertices with a high error value. The algorithm is repeated until a finishing criterion occurs. The details are explained in \cite{Fritzke}.
The principal parameters of the algorithm include:
\begin{itemize}
\item
$N$: $N$ is the number of vertices.
\item
$\epsilon_b$ and $\epsilon_n$: The first closest vertex to the input data and its neighbors are moved towards the input data by fractions
of $\epsilon_b$ and $\epsilon_n$, respectively.
\item
$\alpha_{max}$: The edges older than $\alpha_{max}$ are removed in every step.
\item
$\lambda$: The number of input data used for comparison is $\lambda$.
\item
$\alpha$: The error value of vertices with most error is decreased, after inserting new vertex between them by a multiple of $\alpha$.
\item
$d$: All error variables are decreased in every step by a multiple of $d$.
\end{itemize}

 We tested
different values for GNG parameters and set $N=300 $, $\epsilon_b = 0.05 $, $\epsilon_a = 0.005 $, $\lambda =50$, $\alpha_{max}
=50$, $\alpha =0.5$, and $d =0.995$.

 \subsection{Extracting the outer boundary}
\begin{figure}[htb]
\includegraphics[scale=.9]{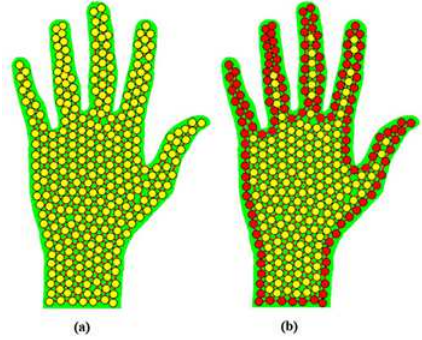}
\centering \caption{\label{fig:GNG and outer} (a) The GNG graph of a hand gesture image, (b) the outer boundary of the GNG graph is specified with red color.}
\end{figure}

The outer boundary of a GNG graph can be computed by an algorithm similar to convex hull.
The algorithm selects the leftmost vertex of the graph and walks clockwise around the graph to reach initial vertices.
More details are described in \cite{mirehi2019hand}.
The GNG boundary is an approximation of the contour of the image so unlike pixel-based boundaries, it is not sensitive to noise on the contour. Figure \ref{fig:GNG and outer} shows an example of a GNG graph and its boundary.

 \subsection{Topological features}
In this section, we introduce the meaningful features that capture the topological and geometrical properties of the graph.
At first, we find the peaks and troughs in the boundary and then defined the features by using them.

 Let $G=(V, E)$ be the GNG and H be the spanning subgraph of G consisting of boundary edges, note that the number of the vertices of G and H is the same. The adjacency matrices of $G$ and $H$ are specified with $A$ and $B$, respectively.
For each peak on the boundary, there are two vertices connecting it to the rest of the image. We select these vertices as the basic vertices of the bulge and the subgraph inside this peak as the bulge itself. The distance of the basic vertices of the bulge in G is less than a multiple $(<1)$ of their distance in H.
\begin{figure}[htb]
\includegraphics[scale=.75]{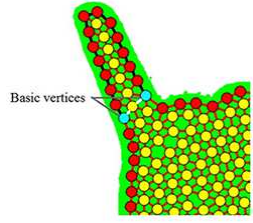}
\centering \caption{\label{fig:bluge} An example of a bulge.}
\end{figure}

 There is a standard for hand and body measurement \cite{klein2012science}. According to this, the length of the middle finger (fingertip to knuckle) is at least 5.5 times its width. The length of the little finger is not smaller than half the length of the middle finger.

 The experimental study indicates that the distance of the basic vertices of a finger is at most 2 in a GNG graph of a hand with 300 vertices, so we consider the distance between the base of one finger is 2, and the length of fingers is greater than 4.
To find the bulges, we compute the matrix $A-B$. Non-zero elements of $A-B$ describes the edges of $G$ that do not belong to the graph $ H $; therefore, $(A-B)^k$ shows the number of walks of length $k$ avoiding $H$ between vertices \cite{Bondy}.

 The pairs of vertices whose corresponding entry in $C=((A-B)^2>0)-((B^3+B^4>0)$ are non-zero candidate as basic vertices of a finger. Furthermore, the pairs of vertices whose corresponding entry in $C=((A-B)^4>0+(A-B)^5>0)-((B^6+B^7+B^8>0)$ are non-zero candidate as basic vertices of sticking fingers.

\begin{figure}[htb]
\includegraphics[scale=.8]{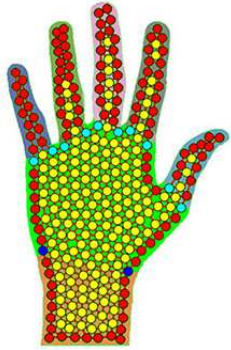}
\centering \caption{\label{fig:gng_bulges}The GNG graph of a hand gesture and its bulges including fingers and wrist.}
\end{figure}

 To find the wrist, we consider the distance between the basic vertices of the wrist in $G$ as 6 or 7. The shape of the wrist is close to a rectangle, so the distance between the basic vertices of the wrist in graph $H$ is at least 11. In a similar way, matrix $$((A -B)^6>0+(A -B)^7>0)- (\sum_{n=8}^{11} B^{n})>0$$ is used for finding the basic vertices of the wrist.
Between candidate pairs of basic vertices to find fingers and wrist, we select the pair with the largest distance in $H$.
Figure \ref{fig:gng_bulges} displays an example of bulges (finger and wrist) from the GNG of a hand gesture. The geometrical and topological features are defined as follows.

\begin{figure*}[htb]
\includegraphics[scale=1.1]{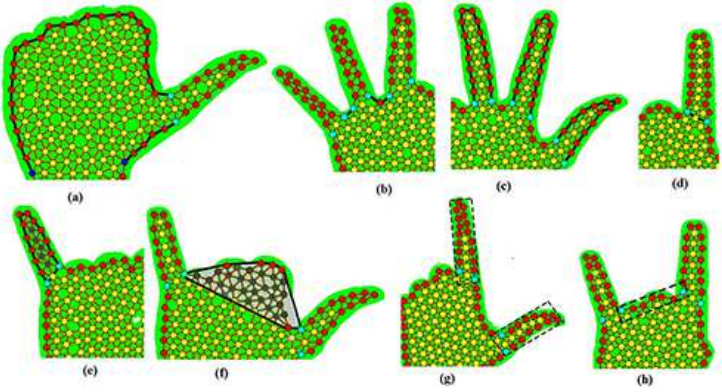}
\centering \caption{\label{fig:new_features} features in a GNG graph of a hand gesture.}
\end{figure*}

\begin{itemize}
\item\textbf{The ratio of distances between fingers and the wrist ($ \mathbf{S_1}$ and $\mathbf{S_2}$)}\\
Feature $S_1$ measures the ratio of distances between the first finger and the wrist, and feature $S_2$ is the ratio of distances between the last finger and the wrist.
There are counterclockwise and clockwise paths from the basic vertices of a finger $i$ to the basic vertices of the wrist in $H$ (We replace other fingers with their base path $P$). The length of the counterclockwise and clockwise paths are specified with $L_1(i)$ and $L_2(i)$, respectively.
Feature $S_1$ for the first finger $i$ is defined as,
$$S_1=\frac{L_1(i)}{L_1(i)+L_2(i)}$$
\\
Feature $S_2$ for the last finger $j$ is defined as,
$$S_2(j)=\frac{L_2(j)}{L_1(j)+L_2(j)}$$

 The features of $S_1$ and $S_2$ illustrate the relative location of the fingers with respect to the wrist.
In Figure \ref{fig:new_features}-a, the paths indicating $L_1$ and $L_2$ for the thumb are shown with black edges.

\item \textbf{Distances between bulges ($\mathbf{D}$)}\\
For consecutive bulges $i$ and $i-1$, $D(i)$ shows the distance between them in $H$ (see Figure \ref{fig:new_features}-b).
% The shortest path between little and thumb in $H$ is specified with gray edges in figure \ref{fig:new_features}.

 \item \textbf{The length of bulges ($\mathbf{R_b}$)}\\
This feature measures the length of a bulge. For a given bulge $i$, $R_b(i)$ is the length of the path between basic vertices of $i$ in $H$.
This path for middle finger, index finger and thumb is specified with black edges in Figure \ref{fig:new_features}-c with lengths 17, 17, and 13, respectively.
%distance between basic vertices of a bulge
%This path is shown with white edges for two fingers in figure \ref{fig:new_features}.

 \item \textbf{The width of bulges ($\mathbf{W_b}$)}\\
This feature contains the distance between the pair of the basic vertices of a bulge (see Figure \ref{fig:new_features}-d).

 \item \textbf{ The number of GNG vertices in a bulge ($\mathbf{N_b}$)}\\
This feature shows the number of GNG vertices that are in a bulge (see Figure \ref{fig:new_features}-e).

 \item \textbf{ The number of GNG vertices that are inside the convex hull of region
between bulges ($\mathbf{N_d}$)}\\
This feature measures the number of vertices inside the convex hull of the shortest path between the two consecutive bulges. In fact, this feature measures the convexity or concavity of the region between the two consecutive bulges. In Figure \ref{fig:new_features}-f, the convex hull of the shortest path between two fingers is shown with black.

 \item \textbf{The aspect ratio of OMBB of a bulge ($\mathbf{O_b}$)}\\
Given a bulge, this feature computes the ratio of the width and length of the oriented minimum bounding box (OMBB) of the bulge.
Figure \ref{fig:new_features}-g shows the OMBB of the index finger and thumb.

 \item \textbf{ The aspect ratio of OMBB of the distance between bulges ($\mathbf{O_d}$)}\\
This feature finds the OMBB of the shortest path between bulges and computes the ratio of the width and length of it (see Figure \ref{fig:new_features}-h).
\end{itemize}

 %We initialize the length of feature vector $F$ with zero value for maximum state that five fingers are open and compute the defined feature $F=[S_1, S_2, D, R, W, N_b, N_d, O_b, O_d]$ for all bulges except the wrist.
% Computed features are arranged in their property in the feature vector. The calculated values are arranged in the feature vector, respectively.

\subsection{Improved Earth Mover\textquotesingle s Distance (IEMD)}
We define a new Earth Mover\textquotesingle s Distance to measure the dissimilarity between the extracted features of the GNG graphs.

 The Earth Mover\textquotesingle s Distance (EMD) is a measure of the distance between two probability distributions over a region \cite{rubner2000earth}.
Different researchers have presented various forms of the EMD in terms of their application \cite{wang2015superpixel,wang2017Liu}.
We compute the IEMD of two GNG graphs as follows.
The first GNG graph is defined as a signature $P =\lbrace(p_{1}, w_{p_1} ), ..., (p_m, w_{p_m})\rbrace$
with m clusters, where $p_i$ represents the cluster, $w_{p_i}$ is the weight of the cluster, and
$Q = \lbrace(q_1, w_{q_1} ), ..., (q_n, w_{q_n})\rbrace$ shows the signature of the second GNG graph with n clusters.
We consider every computed bulge as a cluster. The weight of a bulge is a vector with length 7, which contains the principle defined information of the clockwise order bulges and their relationship with other bulges as follows.
\begin{equation}
\begin{aligned}
%[1.2ex]
w_{p_i}=\begin{cases}[S_2,0,0,0,0,0,0] & \text{i is the wrist} \\[1.2ex]
[S_1,R_b,W_b,N_b,N_d,O_b,O_d] & \text{i is the first bulge} \\[1.2ex]
[D,R_b,W_b,N_b,N_d,O_b,O_d] & \text{otherwise}
\end{cases}
\end{aligned}
\end{equation}

 Since the basic vertices of the wrist are not necessarily located at the wrist of the image and might be on the forearm, we use $S_1$ and $S_2$ instead of $D$ to describe the relative distance of the wrist.

 The maximum number of bulges occurs when all fingers are open. In this case, the number of bulges equals to 6 including five fingers and the wrist. To improve partial matching and reduce mismatching, we insert virtual clusters in signatures such that the number of its clusters equals 6. If m is less than 6, $6-m$ virtual clusters with zero weights be inserted into the signature $P$ and similarly $6-n$ virtual clusters with zero weights be inserted to signature $Q$.
For the two bulges $p_i$ and $p_j$, the cost $c_{ij}$ is defined as

 \begin{equation}
\begin{aligned}
c_{ij}=\begin{cases}
\parallel w_i-w_j \parallel & \text{if i=j} \\[1.2ex]
i\times j\times \parallel w_i-w_j \parallel & \text{if i $\neq$ j} \\[1.2ex]
\end{cases}
\end{aligned}
\end{equation}

 We expect that a good matching preserves the order of bulges.
In this case, the cost is computed as the difference of the weight of these bulges. Otherwise, a penalty in the form $i \times j$ is added to the cost.
We need to find a flow $F = [f_{ij} ]$ between two signatures where $f_{ij} $ is the flow between $p_i$ and $q_j$

%that minimizes the overall cost.
%$WORK(P, Q, F) = \sum_{i=1}^{m} \sum_{j=1}^{n} d_{ij} f_{ij}$
with the following constraints:\\
%$$ f_{ij}\geq 0 ~~ ~~~ 1 \leq i \leq m , ~ 1\leq j \leq n$$
%$$ \sum_{j=1}^{n} f_{ij} \leq w_{p_i} ~~~~1\leq i \leq m $$
%$$ \sum_{i=1}^{m} f_{ij} \leq w_{q_j} ~~~~1\leq j \leq n $$

 %$$ \sum_{i=1}^{m} \sum_{j=1}^{n} f_{ij}= min( \sum_{i=1}^{m}w_{pi}, \sum_{j=1}^{n}w_{qj}) $$

 \begin{equation}
\begin{aligned}
s.t.\begin{cases}

 f_{ij}\geq 0 ~~ ~~~ 1 \leq i \leq m , ~ 1\leq j \leq n \\[1.2ex]
\sum_{j=1}^{n} f_{ij} \leq w_{p_i} ~~~~1\leq i \leq m \\[1.2ex]
\sum_{i=1}^{m} f_{ij} \leq w_{q_j} ~~~~1\leq j \leq n \\[1.2ex]

 \sum_{i=1}^{m} \sum_{j=1}^{n} f_{ij}= min( \sum_{i=1}^{m}w_{pi}, \sum_{j=1}^{n}w_{qj}) \\[1.2ex]
\end{cases}
\end{aligned}
\end{equation}

 Then, IEMD is defined as,
\begin{equation}
IEMD(P, Q) = \dfrac{\sum_{i=1}^{m} \sum_{j=1}^{n} c_{ij} f_{ij}}{\sum_{i=1}^{m} \sum_{j=1}^{n} f_{ij}}
\end{equation}
We used IEMD to measure the dissimilarity between two hand gestures.
This measurement does not depend on the location and depth of the pixels \cite{wang2015superpixel,wang2017Liu}, which results more stability against hand shape variations.
Moreover, IEMD can be applied to measure the dissimilarity in other approaches.
\subsection{Hand gesture recognition}
Finally, we use the k-nearest neighbors (k-NN) algorithm for hand gesture classification. In this algorithm, the value of k is chosen according to the data. In our experiments, the value of k is considered 3 and the
class of a hand gesture is predicted by three of the nearest neighbors of the training class. The size and variety of training sets can affect accuracy.
In order to comprehensively evaluate the proposed approach, the training data sets are chosen as follows.
Using half of the data for training and the other half for testing (h-h), leave-p-subject-out (l-p-o). In (l-p-o) validation protocol, if the dataset includes N subjects, N-p subjects are chosen for training and the remaining p subjects are used for testing. This procedure is repeated for
every combination of p subjects, and then the average accuracy is reported. We choose leave-one-subject-out (l-o-o) and leave-4-subject-out (l-4-o) cross-validation, which are more common protocol to evaluate approaches.

 \section{Experimental study}\label{sec:Experimental study}
In this section, we evaluate and compare our approach with some state-of-the-art approaches such as \cite{ren2011robust}, skeleton matching \cite{ren2013robust}, Hand dominant line \cite{wang2013hand}, H3DF \cite{zhang2013histogram}, VS-LBP \cite{Maqueda2015}, SP-EMD \cite{wang2015superpixel}, CSG-EMD \cite{wang2017Liu}, and GNG+LDA \cite{mirehi2019hand} on different datasets. These datasets are NTU Hand Digits, HKU, HKU multi-angle, and UESTC-ASL datasets. First we introduce these datasets briefly.
% \cite{cheng2016image,Li2018Deep,ren2011robust,Maqueda2015,
\subsection{Datasets}
%\subsection{NTU Hand Digits dataset}
\subsubsection{NTU Hand Digits dataset}
\begin{figure}[htb]
\includegraphics[scale=.8]{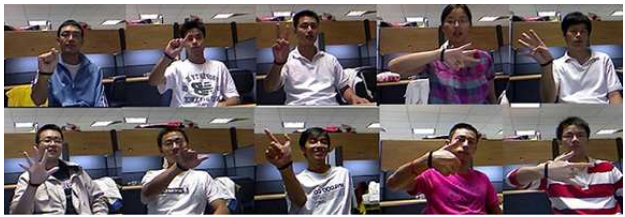}
\centering
\caption{\label{fig:NTU} Some samples from NTU Hand Digits dataset.}
\end{figure}

The NTU Hand Digits dataset is collected with Kinect and includes 1000 color images and their depth maps in cluttered backgrounds.
It contains 10 hand gestures of decimal digits 0-9, which are performed by 10 subjects with 10 samples per gesture.
The subjects pose with variations such as orientation, articulation, and scale in gestures \cite{ren2011robust}. Figure \ref{fig:NTU} shows some of these images.

 \subsubsection{HKU dataset}
\begin{figure}[htb]
\begin{center}
\includegraphics[scale=.8]{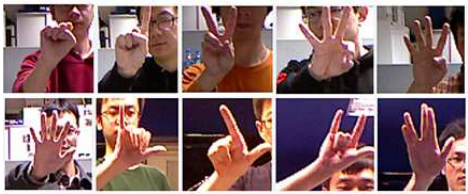}
\end{center}
\centering
\caption{\label{fig:HKU} The gesture samples of HKU dataset. }
\end{figure}

HKU dataset is captured using Kinect from 5 subjects. It consists of 1000 joint color-depth images with 10
gestures from labels 0 to 9. The subjects have performed each gesture in 20 different poses \cite{wang2015superpixel}.
This dataset contains 10 gestures with 20 different poses from 5 subjects. In this dataset, the hand motions include large in-plane rotation and moderate out-of-plane rotation. In Figure, \ref{fig:HKU} gesture samples are shown.

 %\subsection{HKU multi-angle dataset}
\subsubsection{HKU multi-angle dataset}
\begin{figure}[htb]
\includegraphics[scale=1.3]{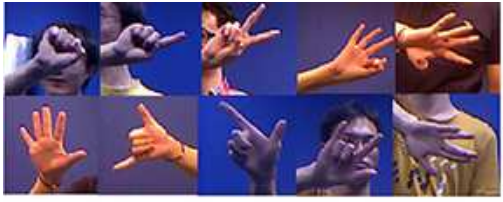}
\centering
\caption{\label{fig:HKU-angle} The gesture samples of HKU multi-angle dataset. }
\end{figure}

The HKU multi-angle hand gesture dataset is an extension of HKU dataset with challenging samples from 4 different viewing angles (approximately 0, 10 and $\pm$ 20) with 5 subjects. The HKU multi-angle includes 2000 color images \footnote{The dataset downloaded from the link reported \cite{wang2015superpixel} contains 2000 images while the authors indicated that the dataset includes 3000 images.} for testing \cite{wang2015superpixel}.
In Figure \ref{fig:HKU-angle}, gesture samples are shown.

 \subsubsection{UESTC-ASL dataset}
\begin{figure}[htb]
\includegraphics[scale=1.1]{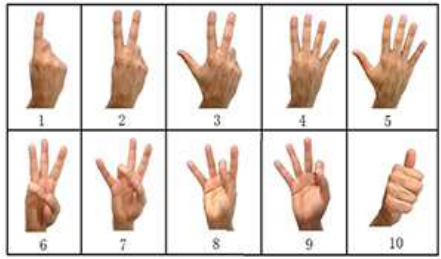}
\centering
\caption{\label{fig:UESTC_ASL} The sample gestures of UESTC-ASL dataset \cite{cheng2016image}. }
\end{figure}

UESTC-ASL dataset is collected 1100 color images from ASL digit gestures by Kinect.
Gestures from 1 to 10 are performed 11 times by 10 subjects in different orientations, depths, and scales \cite{cheng2016image}.
Figure \ref{fig:UESTC_ASL} shows some samples of UESTC-ASL dataset. Due to the high similarity among the gestures of ASL digits and small inter-class variations, this dataset is really challenging.

\subsection{Mean accuracy}
We test the proposed approach on NTU Hand Digits, HKU, HKU multi-angle, and UESTC-ASL datasets in a 3GHz CPU with Matlab implementation. The experimental results and comparison with the state-of-the-art approaches are reported in Tables 1-4.

 Table \ref{table:NTU-2-gesture dataset} shows the results and the comparison on NTU Hand Digits dataset.
The proposed approach is compared with well-known approaches as Thresholding Decomposition \cite{ren2011robust}, skeleton matching \cite{ren2013robust}, Hand dominant line \cite{wang2013hand}, H3DF \cite{zhang2013histogram}, VS-LBP \cite{Maqueda2015}, CSG-EMD \cite{wang2017Liu}, and GNG+LDA \cite{mirehi2019hand}.

\begin{table}[!htb]
\caption{The comparison of performance on NTU Hand Digits dataset}\label{table:NTU-2-gesture dataset}
%\begin{left}
\begin{center}
\scalebox{0.8}{
\begin{tabular}{ l c c}
\hline
\textbf{ Approaches based on (h-h) } & \textbf{ Mean accuracy }\\
\hline
Skeleton matching \cite{ren2013robust} & 78.6 \\
Near-convex Decomposition+FEMD \cite{ren2011robust} & 93.9 \\
Hand dominant line + SVM \cite{wang2013hand} & 97.1 \\
VS-LBP + SVM \cite{Maqueda2015} & 97.3 \\
GNG+LDA \cite{mirehi2019hand} & 98.68\\
GNG-IEMD & \textbf{99.7} \\
\hline
\textbf{ Approaches based on Deep learning} & \textbf{ Mean accuracy}\\
\hline
Deep network + RGB-D images \cite{Li2018Deep} & 98.5\\
\hline
\textbf{ Approaches based on (l-o-o)} & \textbf{ Mean accuracy}\\
\hline
Thresholding Decomposition+FEMD \cite{ren2011robust} & 95\\
Shape context without bending cost \cite{ren2013robust} & 97\\
Shape context with bending cost \cite{ren2013robust} & 95.7\\
Skeleton matching \cite{ren2013robust} & 96 \\
Hand dominant line + SVM \cite{wang2013hand} & 91.1\\
H3DF \cite{zhang2013histogram} & 95.5 \\
%H3DF+SRC \cite{zhang2015histogram} & 97.4 \\
VS-LBP + SVM \cite{Maqueda2015} & 95.9\\
CSG-EMD (shape only) \cite{wang2017Liu} & 99.6\\
CSG-EMD \cite{wang2017Liu} & 99.7\\
GNG+LDA \cite{mirehi2019hand} & 98.6 \\
GNG-IEMD & \textbf{99.9 } \\
\hline
\textbf{ Approaches based on (l-4-o)} & \textbf{ Mean accuracy}\\
\hline
Thresholding Decomposition+FEMD \cite{ren2011robust} & 91.025\\
Shape context without bending cost \cite{ren2013robust} & 92.2\\
Shape context with bending cost \cite{ren2013robust} & 85.375\\
Skeleton matching \cite{ren2013robust} & 90.475 \\
SP-EMD \cite{wang2017Liu} (shap only) & 96.5\\
SP-EMD \cite{wang2017Liu} & 97.2 \\
GNG-IEMD & \textbf{99.3 }\\
%\hline
%\textbf{ Approaches based on (l-9-o)} & \textbf{ Mean accuracy}\\
%\hline
%CSG-EMD (shape only) \cite{wang2017Liu} & 96.2\\
%CSG-EMD \cite{wang2017Liu} & 93\\
%GNG+LDA & \textbf{98} \\
%GNG-IEMD & 95.93 \\

 \hline
%Dynamic Time Warping+FEMD \cite{cheng2016image} & 93.9\\

 \end{tabular}
}
\end{center}
\end{table}

\begin{figure*}
\centering
\begin{subfigure}[b]{0.32\textwidth}
\includegraphics[width=\textwidth]{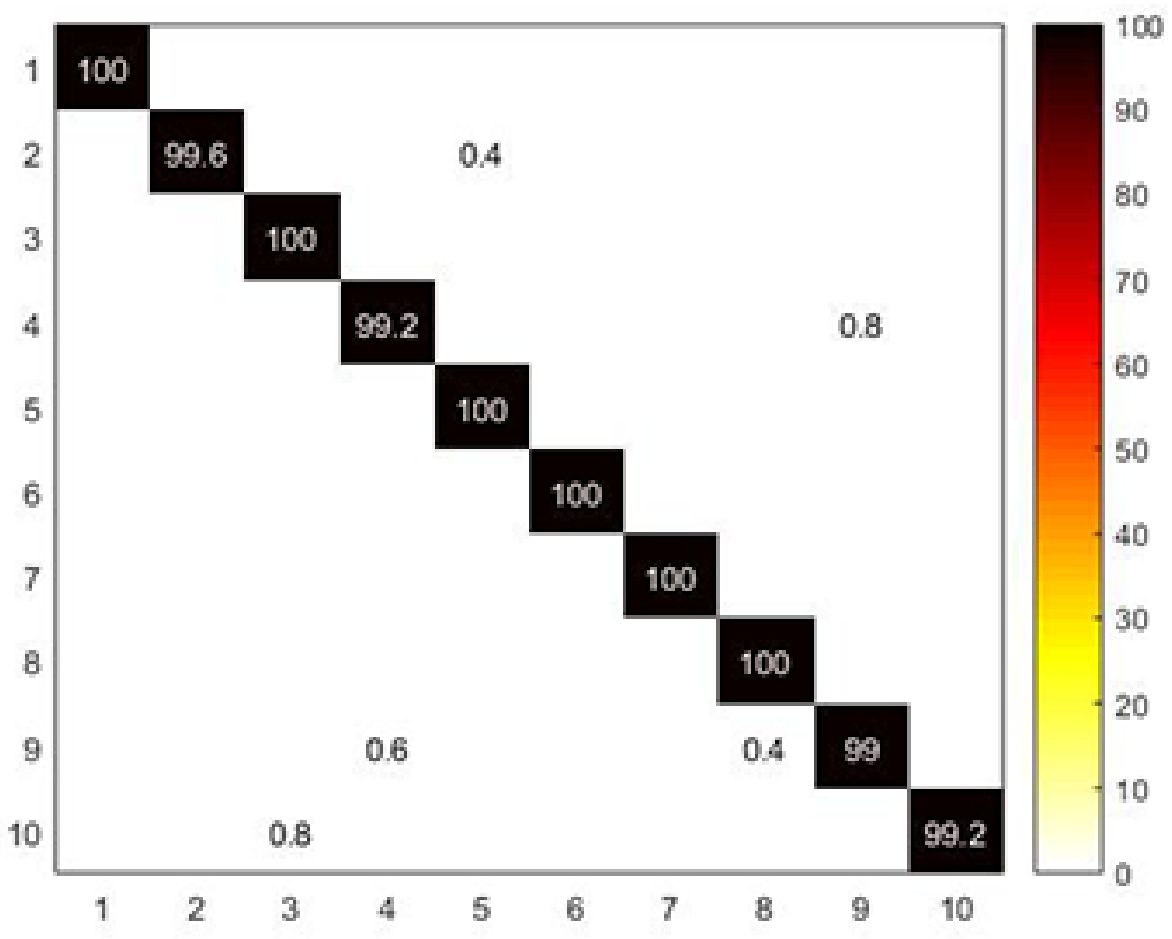}
\caption{NTU dataset (h-h) CV}
\label{fig:Confusion matrix of GNG-IEMD on NTU dataset (h-h) CV}
\end{subfigure}
%\hfill[0.1]
% \qquad
% ~ %add desired spacing between images, e. g. ~, \quad, \qquad, \hfill etc.
%(or a blank line to force the subfigure onto a new line)
% \begin{subfigure}[b]{0.3\textwidth}
% \includegraphics[width=\textwidth]{confison_matrix_NTU_h-h.jpg}
% \caption{A tiger}
%\label{fig:tiger}
%\end{subfigure}
~ %add desired spacing between images, e. g. ~, \quad, \qquad, \hfill etc.
%(or a blank line to force the subfigure onto a new line)
\begin{subfigure}[b]{0.32\textwidth}
\includegraphics[width=\textwidth]{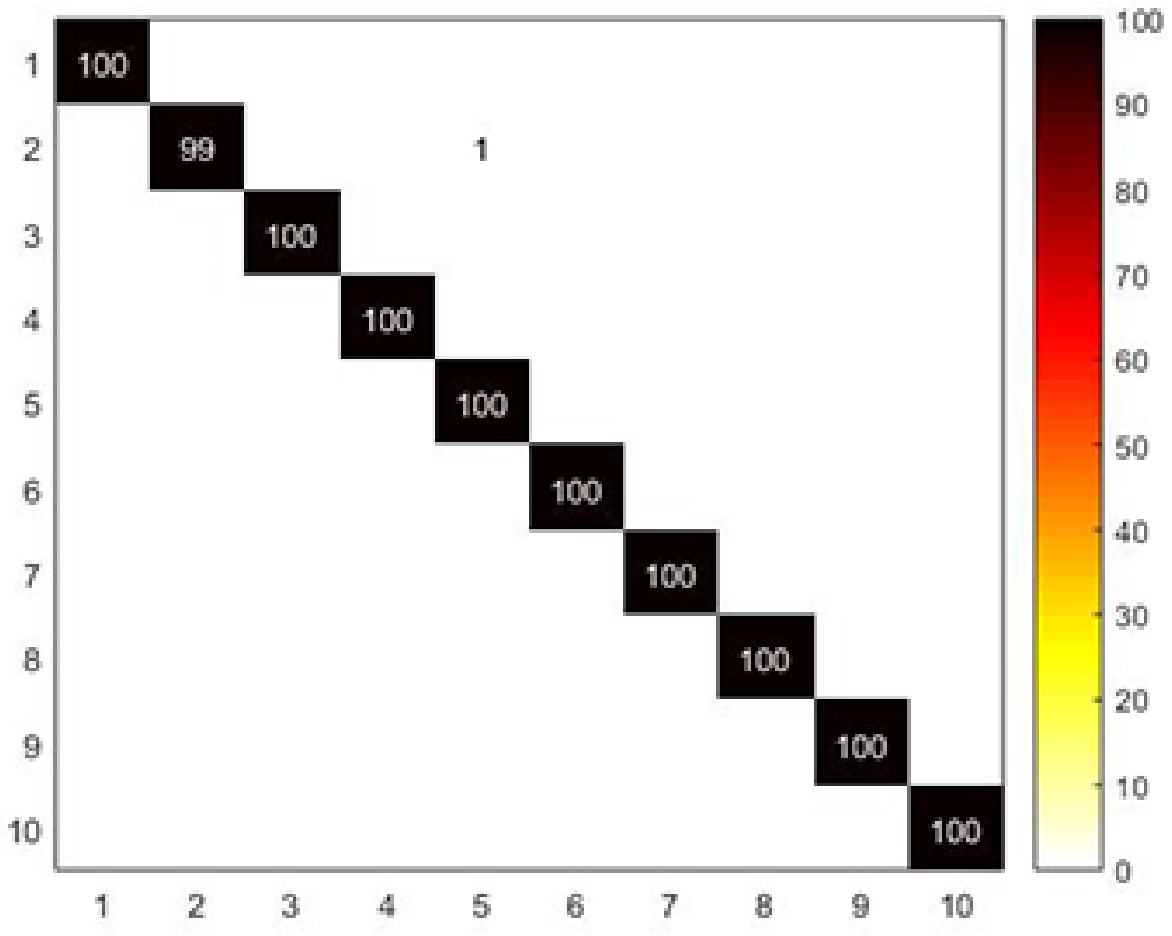}
\caption{ NTU dataset (l-o-o) CV}
\label{fig:Confusion matrix of GNG-IEMD on NTU dataset (l-o-o) CV}
\end{subfigure}
\begin{subfigure}[b]{0.32\textwidth}
\includegraphics[width=\textwidth]{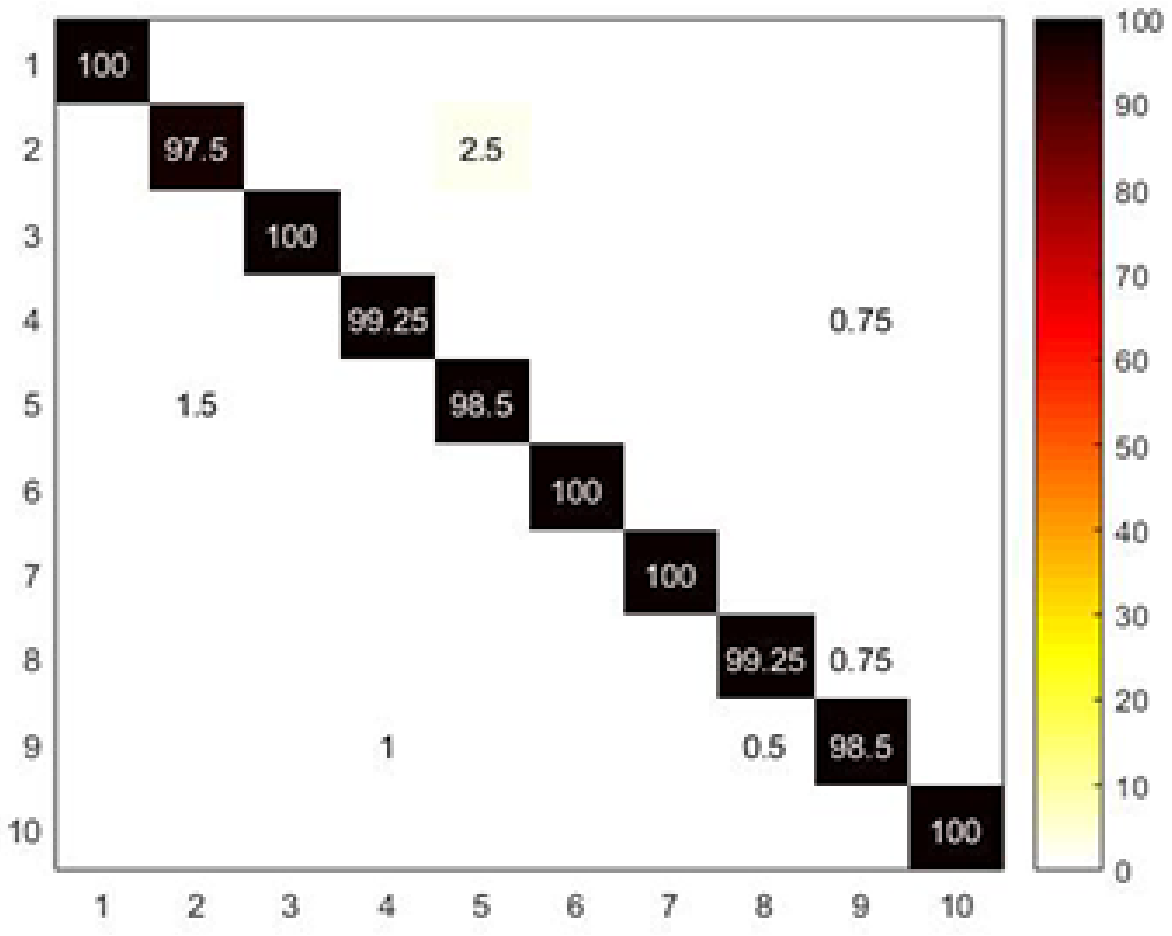}
\caption{ NTU dataset (l-4-o) CV}
\label{fig:Confusion matrix of GNG-IEMD on NTU dataset (l-4-o) CV}
\end{subfigure}

 \begin{subfigure}[b]{0.32\textwidth}
\includegraphics[width=\textwidth]{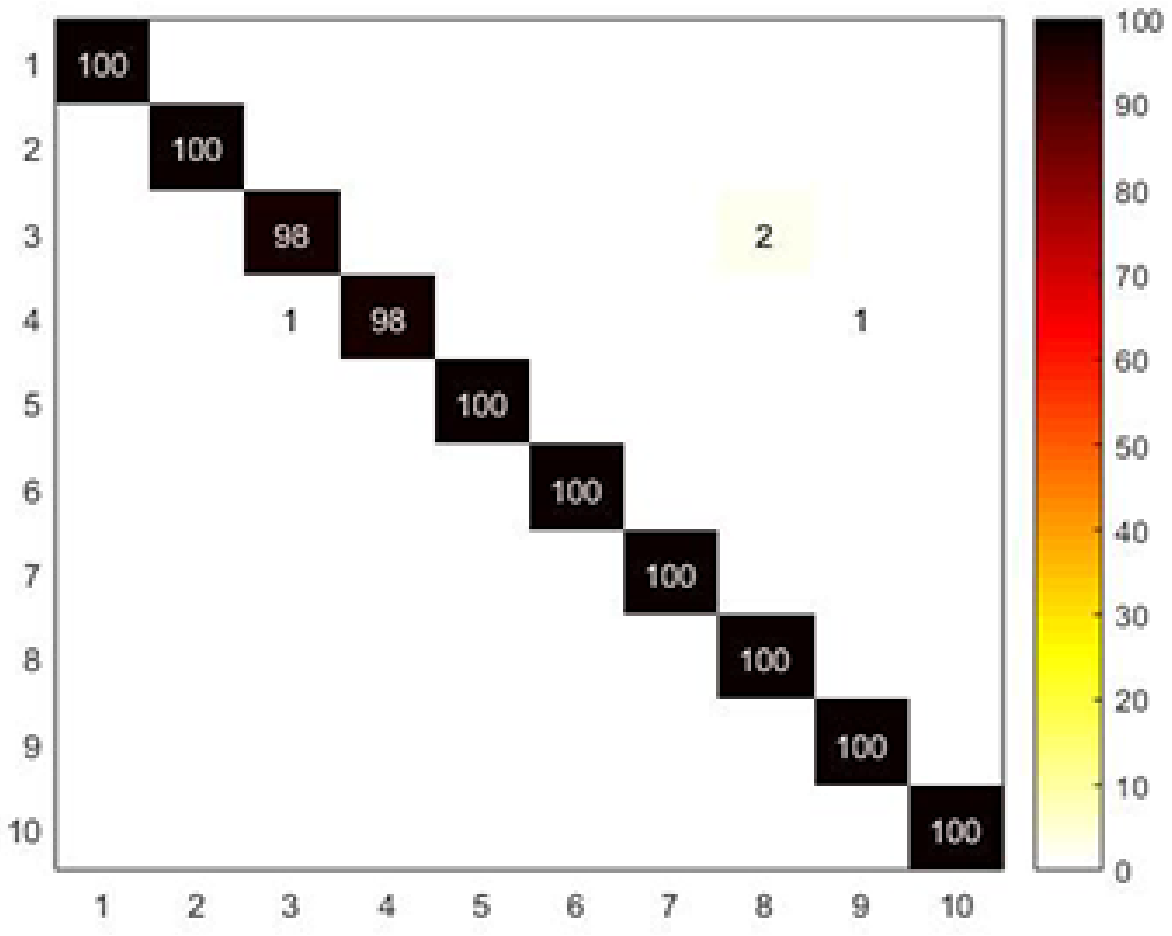}
\caption{ HKU dataset (l-o-o) CV}
\label{fig:Confusion matrix of GNG-IEMD on HKU dataset (l-o-o) CV}
\end{subfigure}
% \vspace{-1}
\begin{subfigure}[b]{0.32\textwidth}
\includegraphics[width=\textwidth]{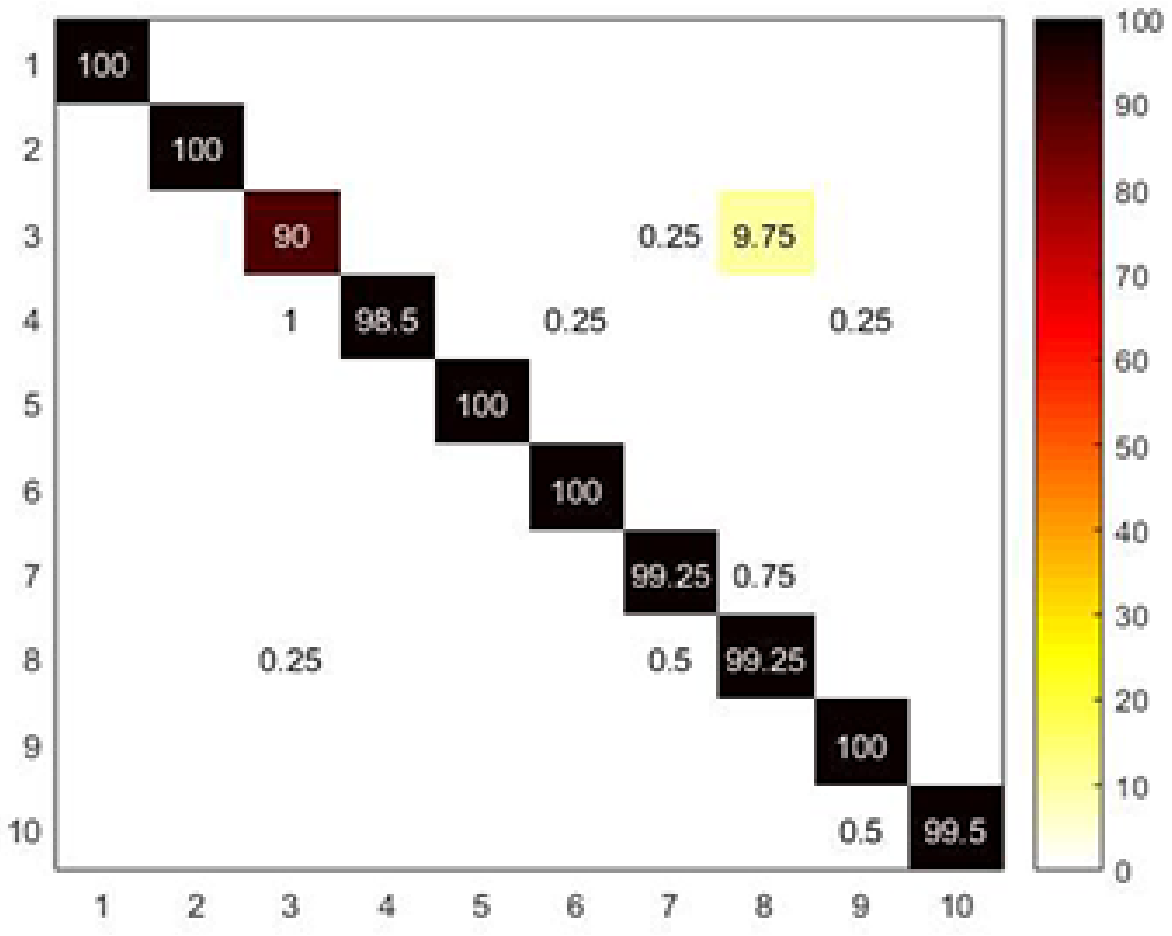}
\caption{HKU dataset (l-4-o) CV}
\label{fig:Confusion matrix of GNG-IEMD on HKU dataset (l-4-o) CV}
\end{subfigure}
\begin{subfigure}[b]{0.32\textwidth}
\includegraphics[width=\textwidth]{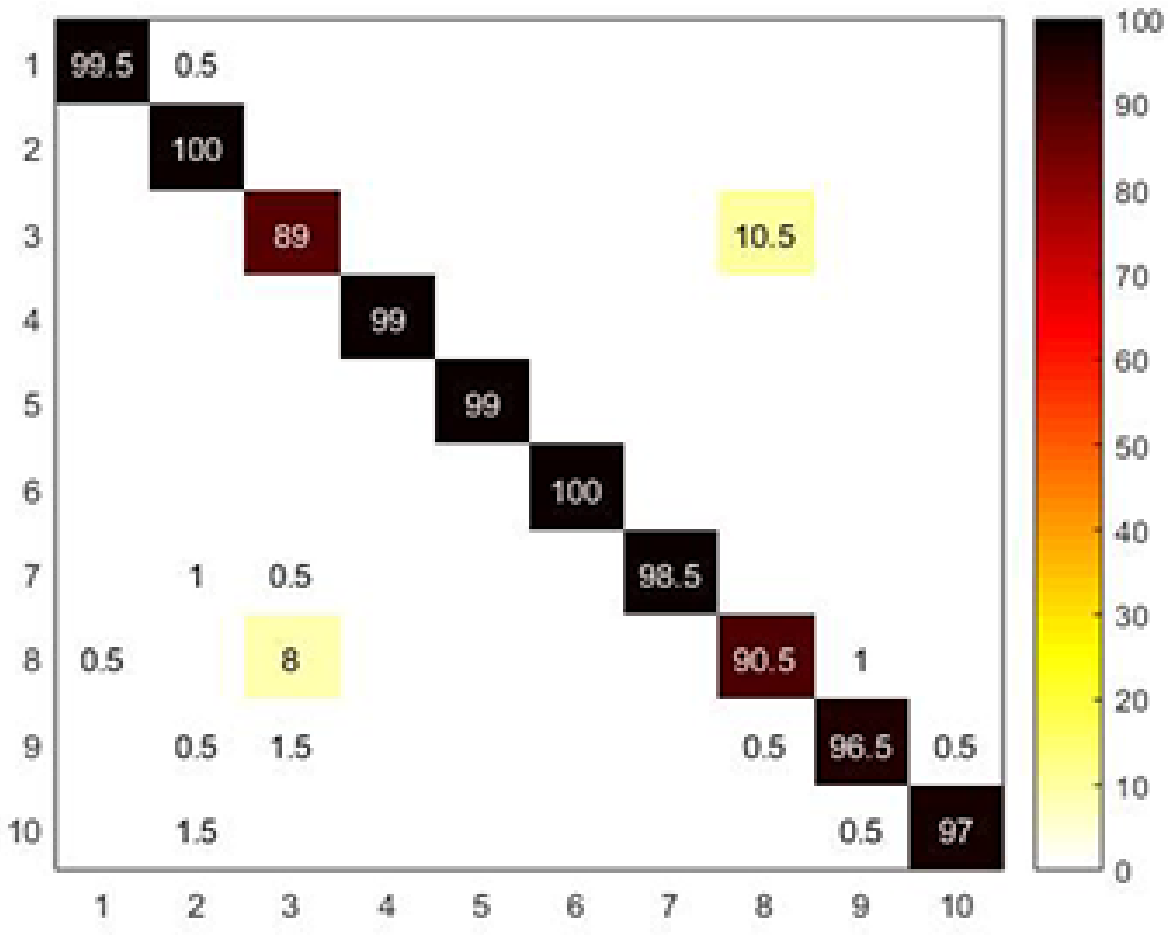}
\caption{HKU multi-angle dataset (l-o-o) CV}
\label{fig:Confusion matrix of GNG-IEMD on HKU multi-angle dataset (l-o-o) CV}
\end{subfigure}
\begin{subfigure}[b]{0.32\textwidth}
\includegraphics[width=\textwidth]{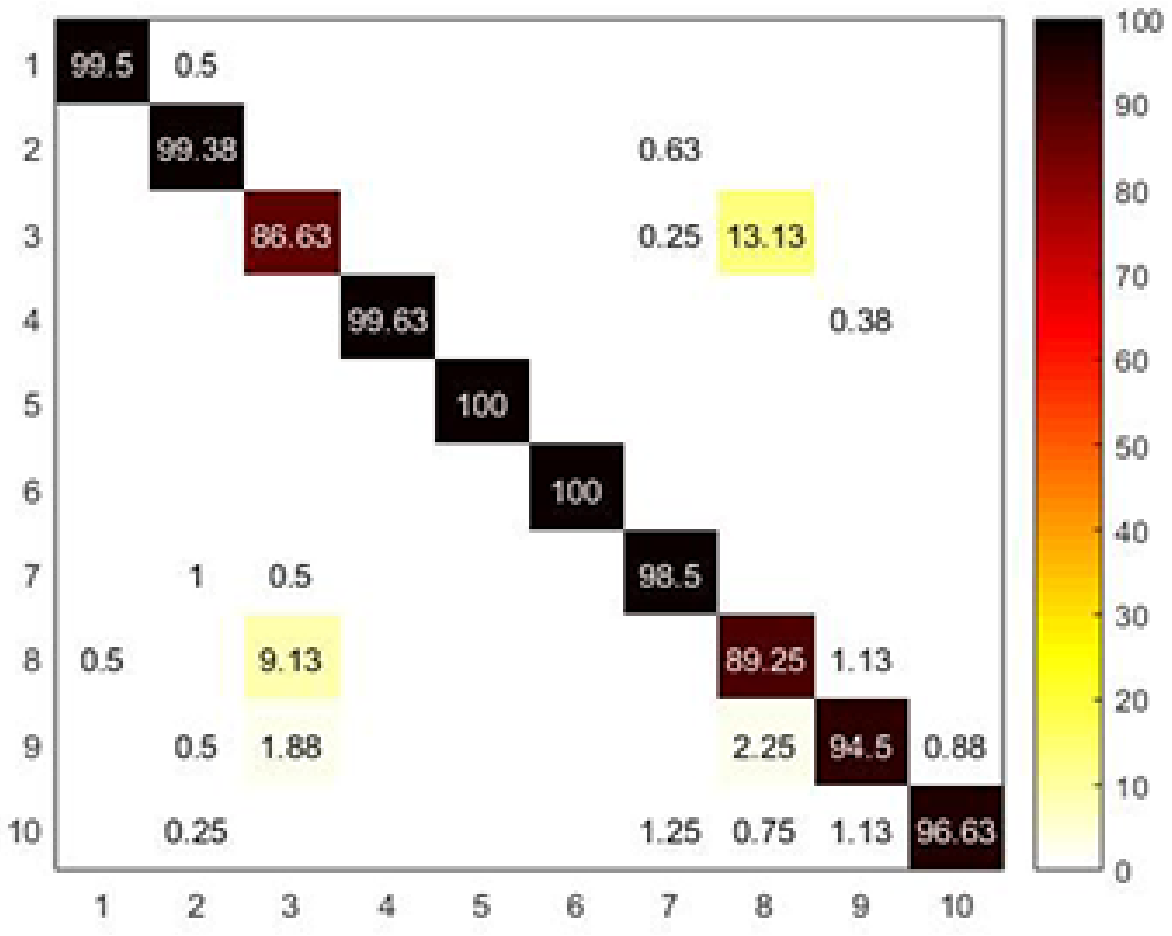}
\caption{HKU multi-angle dataset (l-4-o) CV}
\label{fig:Confusion matrix of GNG-IEMD on HKU multi-angle dataset (l-4-o) CV}
\end{subfigure}
\begin{subfigure}[b]{0.32\textwidth}
\includegraphics[width=\textwidth]{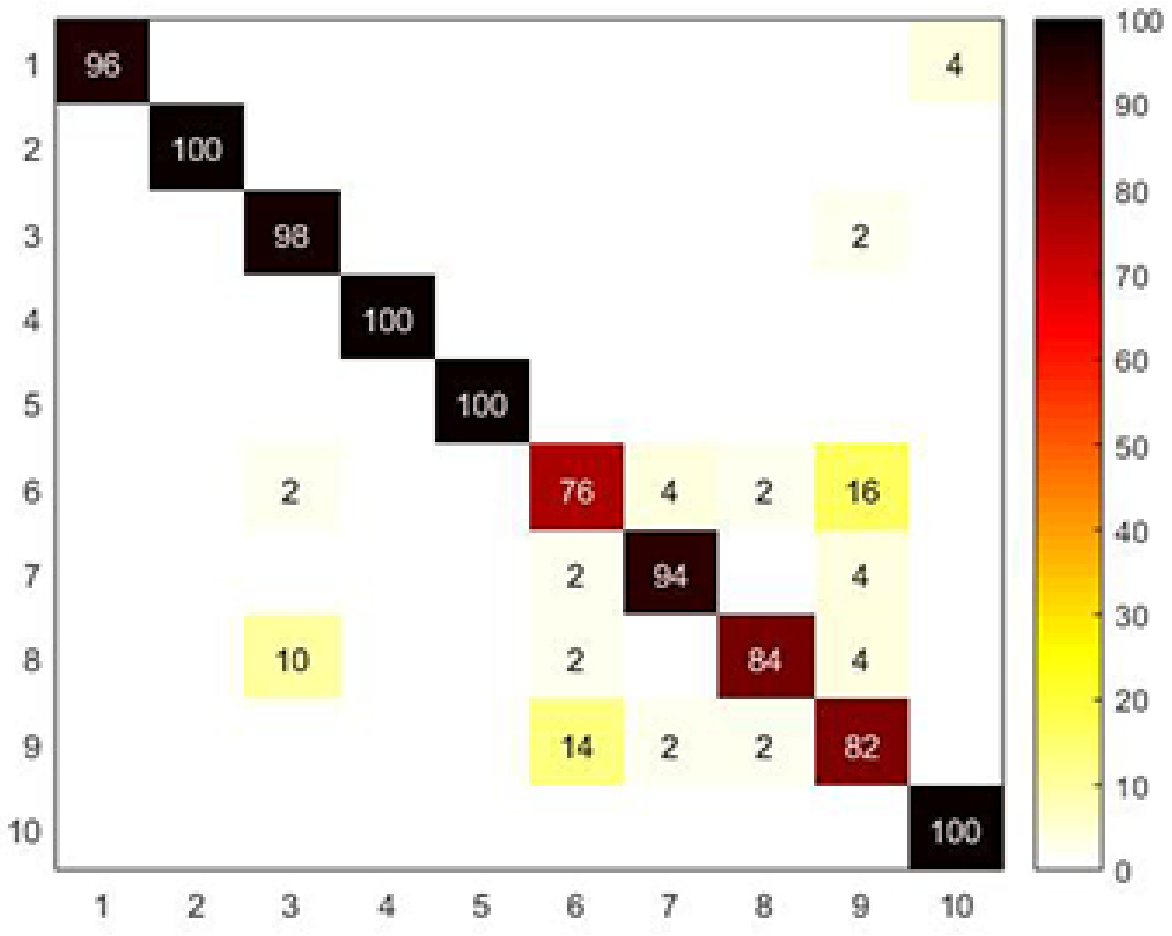}
\caption{UESTC-ASL dataset (h-h).}
\label{fig:Confusion matrix of GNG-IEMD on UESTC-ASL dataset (h-h) CV}
\end{subfigure}
\begin{subfigure}[b]{0.32\textwidth}
\includegraphics[width=\textwidth]{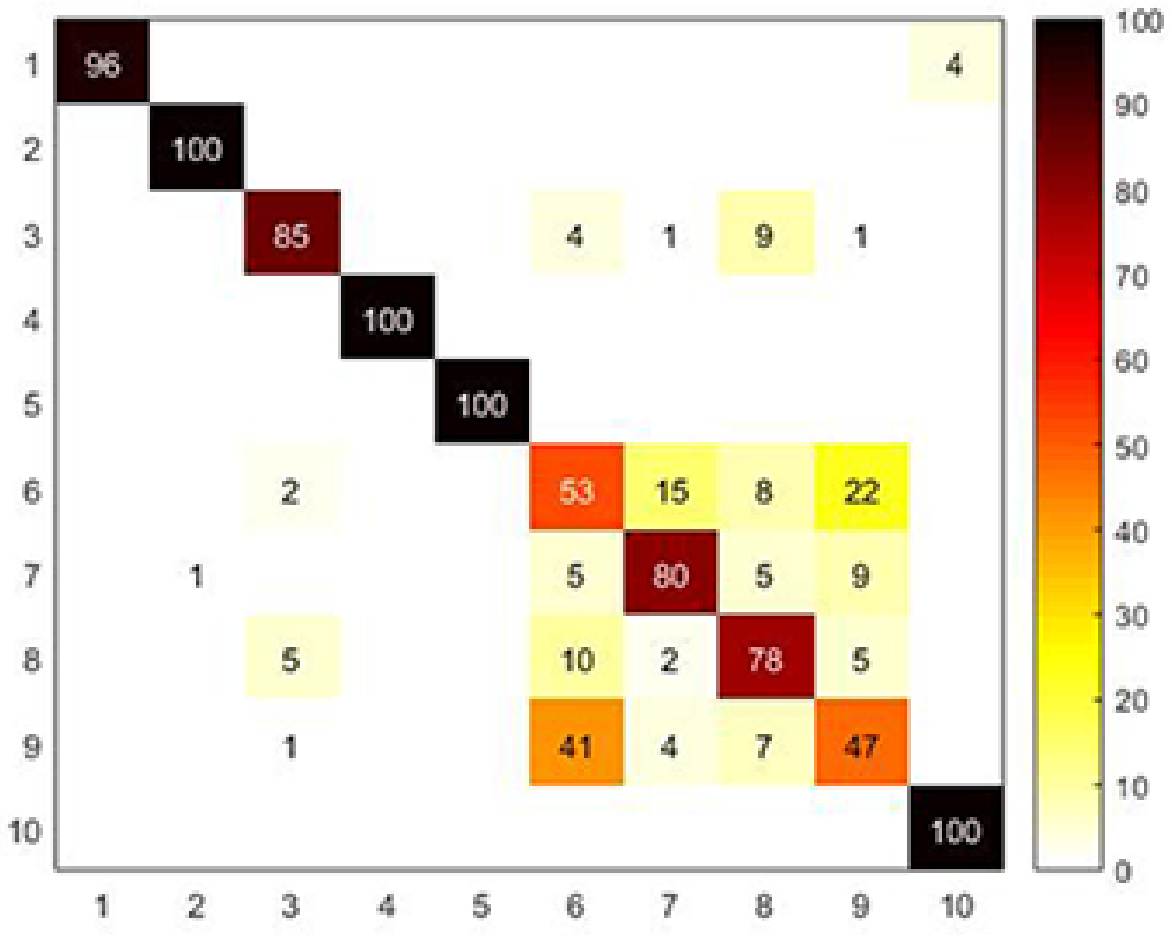}
\caption{UESTC-ASL dataset (I2I).}
\label{fig:Confusion matrix of GNG-IEMD on UESTC-ASL dataset (I2I)}
\end{subfigure}

 \caption{Confusion matrix of GNG+ EMD on NTU, HKU, HKU multi-angle and UESTC-ASL dataset. }\label{fig:Confusion matrix}
\end{figure*}

 As we can see, the proposed approach (GNG-IEMD) achieves the highest mean accuracy of 99.9\%, 99.7\%, and 99.3\% in (h-h), (l-o-o) and (l-4-o) cross validation protocols, respectively.
Although GNG-IEMD uses only the binary images and others utilize the color and depth information \cite{wang2015superpixel,wang2017Liu,Li2018Deep}, our results are more significant.
One of the reasons for this is the use of graph distances instead of Euclidean distances.
Moreover, as presented in Table \ref{table:NTU-2-gesture dataset}, the mean accuracies in (l-4-o) CV and (l-o-o) CV do not differ significantly compared with other approaches, which indicates the insensitivity of the approach to the training data.
Figures \ref{fig:Confusion matrix of GNG-IEMD on NTU dataset (l-o-o) CV}, \ref{fig:Confusion matrix of GNG-IEMD on NTU dataset (l-4-o) CV}, and \ref{fig:Confusion matrix of GNG-IEMD on NTU dataset (h-h) CV} show the confusion matrix of hand gestures in this dataset.
In a few cases, gestures with the same number of fingers have been mismatched. The reason might be the similarity in the topology of gestures and inaccuracy in segmentation.

 The results on HKU dataset is presented in Table \ref{table:HKU-2-gesture hand gesture dataset}.
It can be seen that GNG-IEMD achieved considerable recognition rates in (l-o-o) CV and (l-4-o) CV among other approaches.
Another substantial point is the smallest difference between (l-o-o) and (l-4-o) recognition rates, which results in the independency of our approach on the user and the training data. The confused cases are shown in the Figures \ref{fig:Confusion matrix of GNG-IEMD on HKU dataset (l-4-o) CV}, and \ref{fig:Confusion matrix of GNG-IEMD on HKU dataset (l-o-o) CV}.

 \begin{table}[!htb]
\caption{The comparison of performance on HKU dataset}\label{table:HKU-2-gesture hand gesture dataset}
%\begin{left}
\begin{center}
\scalebox{0.8}{
\begin{tabular}{ l c c c l }
\hline \textbf{ Approaches} & \textbf{(l-o-o CV)} & \textbf{(l-4-o CV)}\\
\hline
Thresholding Decomposition+FEMD \cite{ren2011robust} & 95 & 91 \\
Skeleton matching \cite{ren2013robust} & 96 & 90.5 \\
SP-EMD (shape only) \cite{wang2015superpixel} & 98.7 & 96.1 \\
SP-EMD \cite{wang2015superpixel} & 99.2 & 97.3 \\
CSG-EMD (shape only) \cite{wang2017Liu} & 99.4 & 97.4 \\
CSG-EMD \cite{wang2017Liu} & 99.4 & 97.9 \\
%GNG+LDA & & \\
GNG-IEMD & \textbf{99.6 } & \textbf{98.65}\\
\hline
\end{tabular}
}
\end{center}
\end{table}

 \begin{table}[!htb]
\caption{The comparison of performance on HKU multi-angle hand gesture dataset}\label{table:HKU multi-angle hand gesture dataset}
%\begin{left}
\begin{center}
\scalebox{0.8}{
\begin{tabular}{ l c c c l }
\hline \textbf{ Approaches} & \textbf{(l-o-o CV)} & \textbf{(l-4-o CV)}\\
\hline
Thresholding Decomposition+FEMD \cite{ren2011robust} & 96.2 & 89.7 \\
Skeleton matching \cite{ren2013robust} & 95.1 & 90.3 \\
SP-EMD (shape only) \cite{wang2015superpixel} & 95.3 & 92.5 \\
SP-EMD \cite{wang2015superpixel} & 97.8 & 94.7 \\
CSG-EMD (shape only) \cite{wang2017Liu} & 96.1 & 93.7 \\
CSG-EMD \cite{wang2017Liu} & 97.9 & 95.6 \\
%GNG+LDA & & \\
GNG-IEMD & 96.9 & \textbf{ 96.4}\\
\hline
\end{tabular}
}
\end{center}
\end{table}
The evaluation on HKU multi-angle dataset is shown in Table \ref{table:HKU multi-angle hand gesture dataset}. The recognition accuracies of the proposed approach in both (l-o-o) and (l-o-4) CV are appropriate, which shows the stability of our approach against rotation. Not that GNG-IEMD does not use depth and color data for recognition. Figures \ref{fig:Confusion matrix of GNG-IEMD on HKU multi-angle dataset (l-o-o) CV} and \ref{fig:Confusion matrix of GNG-IEMD on HKU multi-angle dataset (l-4-o) CV} show the confusion matrix.

 \begin{table*}[!htb]
\begin{center}
\caption{The performance on UESTC-ASL dataset}\label{table:UESTC-ASL}
\scalebox{0.75}{
\begin{tabular}{ | c | c c c c c| }
\hline Input sign & Thresholding Decomposition+FEMD \cite{ren2011robust} & I2I-DTW \cite{cheng2016image} & I2C-DTW \cite{cheng2016image} & GNG-IEMD ( I2I) & GNG-IEMD (h-h)\\
\hline
1 & 100 & 92 & 100 & 96 & 96 \\
2 & 100 & 99 & 100 & 100 & 100\\
3 & 95 & 99 & 95 & 85 & 98\\
4 & 100 & 96 & 100 & 100 & 100 \\
5 & 100 & 98 & 100 &100 & 100\\
6 & 59 & 58 & 80 & 53 & 76\\
7 & 80 & 66 & 77 & 80 & 94 \\
8 & 64 & 73 & 90 & 78 & 84\\
9 & 57 & 44 & 72 & 47 & 82 \\
10 & 73 & 93 & 91 & 100 &100\\

 \hline Average & 82.8 & 81.8 & 90.5& \textbf{83.9} & \textbf{93}\\
\hline
\end{tabular}
}
\end{center}
\end{table*}

 The number of studies in UESTC-ASL dataset is limited \cite{ren2011robust,cheng2016image}.
We compared GNG-IEMD with these approaches in Table \ref{table:UESTC-ASL}.
In \cite{cheng2016image}, two Dynamic Time Warping approaches including I2I-DTW, I2C-DTW and, FEMD \cite{ren2011robust} were evaluated on UESTC-ASL.
In Image-to-Image Dynamic Time Warping (I2I-DTW) approach, the distance between the testing sample and all training samples is computed, while Image-to-Class Dynamic Time Warping (I2C-DTW) approach searches for the minimal
warping path between a test sample and a training sample's compositional features \cite
{cheng2016image}.
We evaluated the proposed approach on UESTC-ASL in (I2I) and (h-h).
In (I2I), we choose randomly one image of each subject for training and the rest images for testing.
%Our approach achieves the high accuracies of 83.9\% and 93\%, while Thresholding Decomposition+FEMD, (I2I-DTW), and (I2C-DTW) achieved 82.8\%, 81.8\%, and 90.5\%, respectively. The confused cases can be seen in Figures \ref{fig:Confusion matrix of GNG-IEMD on UESTC-ASL dataset (I2I)} and \ref{fig:Confusion matrix of GNG-IEMD on UESTC-ASL dataset (h-h) CV}.
Our approach achieves the high accuracy of 83.9\% in (I2I), while Thresholding Decomposition+FEMD, and (DTW) achieved 82.8\%, and 81.8\%, respectively. Also, our result in (h-h) is 93\%, which is considerable. The mismatched cases can be seen in Figures \ref{fig:Confusion matrix of GNG-IEMD on UESTC-ASL dataset (I2I)} and \ref{fig:Confusion matrix of GNG-IEMD on UESTC-ASL dataset (h-h) CV}.

 Because of the small inter-class variations in UESTC-ASL dataset, the hand gestures look very similar at some viewpoints.
Especially, gestures with 3 fingers are similar, such as gestures 6 and 9, and also gestures 3 and 7.

 \subsection{Sensitivity Analysis}
The variety in the shape of fingers and hand poses has made hand shape variation problem a significant challenge.
Researchers introduced shape-based features to address this issue. We introduce graph-based features that can extract stable topological features in case of these variations.

 We used the GNG graphs that do not depend on the size of the image, so the features are independent of scale. Since graphs are robust with respect to rotation and articulation, the properties of the GNG graphs were not influenced by rotation and articulation. In \cite{mirehi2019hand}, different experiments were performed to evaluate GNG graphs against scale and rotation, and the results proved the stability of GNG graphs in these cases.
The outer boundary of the GNG graphs is a coarse estimation of the boundary of the object. It describes the overall shape of the object without high dependency on the boundary pixels of the object. This leads to improvement in stability against noise, which is an unavoidable and challenging problem in hand gesture recognition.
Although using sample thresholding for segmentation results images with lots of boundary noises, our approach achieves the higher recognition rate compared with the state-of-the-art. Sensitivity to noise and GNG parameters was evaluated in \cite{mirehi2019hand}.

 \section{Conclusion}\label{sec:conclusion}
In this paper, we proposed a new graph-based approach for hand gesture recognition (GNG-IEMD) with less dependency on pixels compared to other existing approaches.
The hand image is modeled by a GNG graph and the topological and geometrical features of the graph are extracted, and then the dissimilarity of the hand gestures is measured by Improved Earth Mover\textquotesingle s Distance.
In hand gesture recognition, both the boundary and interior information are utilized.
The boundary of the GNG graph models the contour of the image and shows the overall shape of the hand. Hence, the proposed approach is not sensitive to noise on the contour.
To test the performance of GNG-IEMD experimentally, we selected 4 known real-life datasets NTU Hand Digits,
HKU, HKU multi-angle, and UESTC-ASL of ASL Digits. We applied GNG-IEMD on them and compared the results with the state-of-the-art. Our result shows the higher performance of our approach.

 % \section*{Conflict of interest}
%
% The authors declare that they have no conflict of interest.

 % BibTeX users please use one of
 % \bibliographystyle{plain}
 %\bibliographystyle{spbasic} % basic style, author-year citations
%\bibliographystyle{spmpsci} % mathematics and physical sciences
%\bibliographystyle{spbasic}
%\bibliographystyle{splncs}
%\bibliographystyle{spphys} % APS-like style for physics
%\bibliography{} % name your BibTeX data base
%\bibliographystyle{plain}
%\bibliographystyle{unsrtnat}
%\bibliographystyle{unsrt}
% Non-BibTeX users please use
%\bibliography{ref}

\end{document}